\PassOptionsToPackage{table}{xcolor}
\documentclass[journal=jacsat,table]{achemso}
\usepackage[T1]{fontenc}
\usepackage[utf8]{inputenc}
\usepackage{booktabs}
\usepackage{amsmath, amssymb}
\usepackage{algorithm}
\usepackage{algpseudocode}
\usepackage{bm}
\usepackage{geometry}
\usepackage{caption}
\usepackage{array}
\usepackage{tabularx}
\usepackage{graphicx}
\usepackage{float}
\usepackage{subcaption}
\usepackage{multirow}
\usepackage[table]{xcolor}
\usepackage{hyperref}
\usepackage{orcidlink}
\usepackage[version=3]{mhchem}
\usepackage{siunitx}
\usepackage{indentfirst}
\author{XiaYu Liu}
\affiliation[University of Electronic Science and Technology of China]
{School of Mathematical Sciences, University of Electronic Science and Technology of China, Chengdu 610054, China}
\author{Chao Fan}
\affiliation[Chengdu University of Technology]
{College of Management Science, Chengdu University of Technology, Chengdu 610059, China}
\author{Yang Liu}
\affiliation[University of Electronic Science and Technology of China]
{School of Mathematical Sciences, University of Electronic Science and Technology of China, Chengdu 610054, China}
\author{Hou-biao Li}
\affiliation[University of Electronic Science and Technology of China]
{School of Mathematical Sciences, University of Electronic Science and Technology of China, Chengdu 610054, China}
\email{lihoubiao0189@163.com}
\title[An \textsf{achemso} demo]
{Multi-Level Fusion Graph Neural Network for Molecule Property Prediction}
\abbreviations{IR,NMR,UV}
\begin{document}
\begin{abstract}
Accurate prediction of molecular properties is essential in drug discovery and related fields. However, existing graph neural networks (GNNs) often struggle to simultaneously capture both local and global molecular structures. In this work, we propose a Multi-Level Fusion Graph Neural Network (MLFGNN) that integrates Graph Attention Networks and a novel Graph Transformer to jointly model local and global dependencies. In addition, we incorporate molecular fingerprints as a complementary modality and introduce a mechanism of interaction between attention to adaptively fuse information across representations. Extensive experiments on multiple benchmark datasets demonstrate that MLFGNN consistently outperforms state-of-the-art methods in both classification and regression tasks. Interpretability analysis further reveals that the model effectively captures task-relevant chemical patterns, supporting the usefulness of multi-level and multi-modal fusion in molecular representation learning.
\end{abstract}
\section{Introduction}
Molecular property prediction is a fundamental task in chem-informatics and plays a crucial role in various stages of drug discovery, including virtual screening\cite{luttens2025rapid}, lead optimization\cite{gusev2023active}, and toxicity assessment\cite{tan2023hi}. Accurately predicting physicochemical and biological properties of molecules can significantly reduce the cost and time associated with experimental validation and clinical trials\cite{pan2021molgpka}. Traditionally, this task relied heavily on handcrafted features and rule-based systems, which often struggled to generalize across diverse chemical spaces\cite{jiang2021could}. With the growing availability of molecular data and the advancement of computational methods, there has been increasing interest in data-driven approaches, particularly those based on machine learning, to model complex structure-property relationships more effectively\cite{rong2020self}.\par
In recent years, deep learning (DL) has achieved remarkable success in a variety of domains, such as computer vision \cite{krizhevsky2012imagenet} and natural language processing \cite{hirschberg2015advances}, largely driven by improvements in model architectures and the availability of large-scale data. These advances have inspired a wave of applications in scientific and biomedical fields, including atomic charge prediction \cite{wang2021deepatomiccharge}, molecular property modeling \cite{guo2021few}, chemical reaction forecasting \cite{li2023deep}, molecular generation \cite{jin2018junction} and drug property prediction\cite{zhu2023associative,zhu2023drug,zhu2024drug,zhu2025drug}. In particular, deep learning has shown great promise in molecular property prediction by learning generalizable representations directly from raw molecular data, thereby overcoming the limitations of handcrafted descriptors.\par
Among deep learning methods, graph neural networks (GNNs) have become the dominant paradigm for molecular modeling due to their ability to operate directly on molecular graph structures, where atoms are represented as nodes and bonds as edges. A variety of GNN architectures have been proposed for this task \cite{kipf2016semi,gilmer2017neural,hamilton2017inductive,xu2018powerful}.
However, many of these architectures struggle to effectively integrate both local chemical environments and global molecular context, limiting their capacity to capture complex structure-property relationships. One prominent example of such complexity is the activity cliff phenomenon, in which small structural modifications, such as the addition, removal, or substitution of a single atom or function group, can result in disproportionately large changes in biological activity. Accurately modeling activity cliffs requires molecular representations that are both locally sensitive and globally aware, which many existing GNN-based models fail to achieve. To address this limitation, many researches have explored pretraining strategies to enhance the ability of GNNs to capture both local and global molecular information\cite{qiao2025self,zhang2024pre,chen2025pretraining,wang2022molecular}. However, pretraining approaches may suffer from negative transfer, where the knowledge learned during pretraining adversely affects performance on downstream tasks, thereby limiting their effectiveness. In light of these limitations, an alternative line of research focuses on architectural innovations that can inherently balance local and global molecular information without relying heavily on pretraining.\par
In response to these challenges, we design a unified framework that simultaneously address two key aspects of molecular representation: the fusion of local and global structural information, and the incorporation of chemically meaningful features from complementary modalities. While GATs\cite{velickovic2017graph,brody2021attentive} and Graph Transformers\cite{yun2019graph,maziarka2020molecule,maziarka2024relative} have shown promise individually, their combination offers a more comprehensive view of molecular graphs by capturing both neighborhood-level patterns and long-range dependencies.\par
To further enhance molecular representation, we incorporate molecular fingerprints as an additional modality, motivated by their ability to encode domain knowledge in the form of predefined substructures, such as functional groups, ring systems, and pharmacophores\cite{cai2022fp,teng2023molfpg}. These features offer strong chemical priors that are particularly useful for capturing subtle structural motifs, especially in scenarios like activity cliffs where small local changes can lead to significant functional differences. Moreover, integrating fingerprints with graph-based representations can compensate for each other's limitations: while molecular graphs provide flexible and learnable topology-aware embeddings, fingerprints contribute fixed, interpretable descriptors that may be more robust to noise in graph structure or representation sparsity. Previous studies have demonstrated the benefits of such multi-modal integration\cite{jiang2022multigran,zhang2021motif}, but have largely overlooked the importance of selectively filtering task-irrelevant or redundant fingerprint features, which may degrade performance if naively combined\cite{cai2022fp,teng2023molfpg}
. To address this, we introduce a cross-attention mechanism that adaptively aggregates and filters information from both molecular graphs and fingerprints, ensuring that only task-relevant signals contribute to the final representation.\par
Building upon these insights, we propose the Multi-Level Fusion Graph Neural Network(MLFGNN) for molecular property prediction, which performs both intra-graph(local-global) and inter-modal(graph-fingerprint) fusion. The graph branch integrates GATs and a newly designed graph Transformer to effectively capture hierarchical structural information within molecular graphs. In parallel, the fingerprint branch incorporates domain-informed substructure features that provide complementary chemical priors. To harmonize these two modalities, a final cross-attention layer adaptive aggregates and filters their representations, allowing the model to focus on task-relevant features. Extensive experiments on multiple benchmark datasets demonstrate that MLFGNN achieves strong predictive performance while offering improved interpretability and generalization across diverse molecular tasks.\par
    \textbf{Our contributions can be summarized as follows:}
\begin{itemize}
    \item Introduction of a novel Graph Transformer architecture integrated with Graph Attention Networks(GATs), enabling more effective fusion of local and global molecular representations.
    \item A cross-modal fusion strategy combining graph-based and fingerprint-based features using adaptive attention.
    \item Empirical validation through ablation and visualization studies demonstrates the model's generalizability and interpretability, highlighting its focus on chemically relevant substructures.
\end{itemize}
\section{MATERIALS AND METHODS}
\subsection{Problem statement}
SMILES is a textual representation of molecular structures that encodes atoms and bonds as ASCII strings. Each SMILES string $x_{i}$ can be converted into a molecular graph $G_{i}=(\mathcal{V}_{i},\mathcal{E}_{i})$, where
$\mathcal{V}_{i}$ is the set of atoms and $\mathcal{E}_{i}$ is the set of chemical bonds. Given a data set $D=\{(x_{i},y_{i})|x_{i}\in X,y_{i}\in P\}$, where $X$ denotes the set of SMILES strings and $P$ represents the corresponding molecular properties, our aim is to predict molecular properties from structure. Let $M=\{G|G_{i}\in G\}$ denote the set of molecular graphs derived from the dataset. The objective is to leverage both the SMILES representations and their corresponding molecular graphs to predict the associated molecular properties. To achieve this objective, we employ deep learning techniques to train model parameters by minimizing a predefined loss function, thereby enabling more accurate prediction of molecular properties.
\subsection{The architecture of MLFGNN}
\begin{figure}[H]
    \centering
    \includegraphics[width=\textwidth]{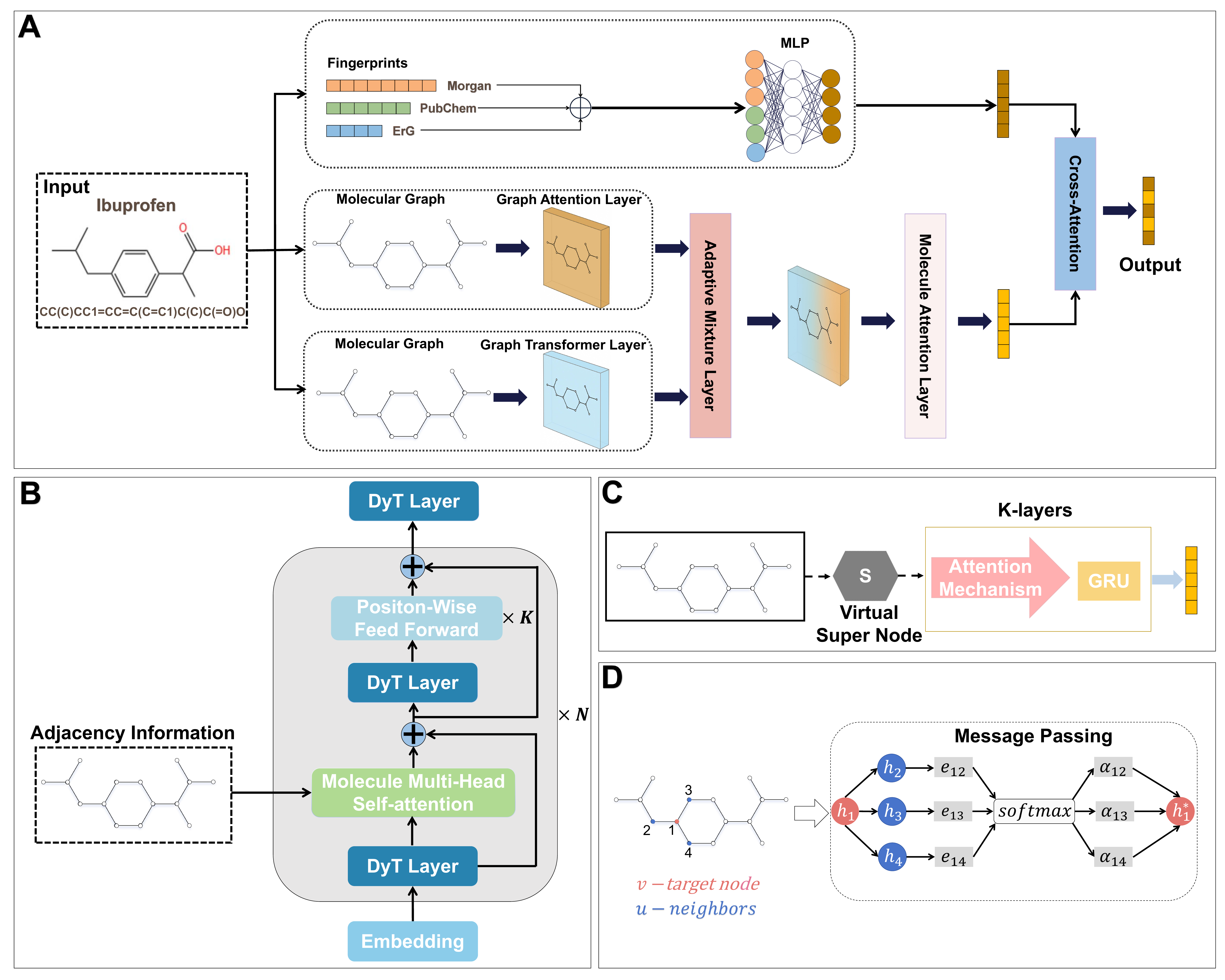}
    \caption{The Multi-Level Fusion Graph Neural Network Architecture (A) The overview of model. (B) The architecture Graph Tansformer Layer. (C)The details of Molecule Attention Layer (D)The Message Passing process of Graph Attention Layer.}
    \label{fig:Architecture}
\end{figure}

\begin{algorithm}
\caption{The Multi-Level Fusion Graph Neural Network}
\begin{algorithmic}[1]
\Statex \hspace*{-\algorithmicindent} \textbf{Input:} SMILES string $S$, Layer $L_{1}$, Layer $L_{2}$
\Statex \hspace*{-\algorithmicindent} \textbf{Output:} Final prediction result $\hat{y}$.
\State $M_{i}=Morgan(S),P_{i}=PubChem(S),E_{i}=PharmaERG(S)$
\State $U_{i}=[M_{i}\text{ }||\text{ }P_{i}\text{ }||\text{ }E_{i}]$
\State $FP_{i}=MLP(U_{i})$
\State Molecular Graph $G=(\mathcal{V},\mathcal{E})=GraphFunction(S)$
\State $V_{i}=InitialNode(v_{i}),E_{uv}=InitialEdge(e_{uv}),v_{i}\in \mathcal{V},e_{vu}\in \mathcal{E}$
\For{$l = 1$ to $L_{1}$}
  \For{$v_{i} \in \mathcal{V}$}
    \If{$l==1$}
    \State $g_{T}=GraphTransformer(V_{i})$
    \Else
    \State $g_{T}=GraphTransformer(g_{T})$
    \EndIf
  \EndFor
\EndFor
\For{$l=1$ to $L_{2}$}
    \For{$v_{i}\in \mathcal{V},e_{iu}\in Neightbor(v_{i})$}
    \If{$l==1$}
    \State $g_{A}=GraphAttention(V_{i},E_{iu})$
    \Else
    \State $g_{A}=GraphAttention(g_{A})$
    \EndIf
    \EndFor
\EndFor
\State$G_{M}=Mixture(g_{A},g_{T})$
\State$G_{M}=MoleculeAttention(G_{M})$
\State $F_{M}=CrossAttention(FP_{i},G_{M})$
\State \textbf{return} $\hat{y} = \text{MLP}(F_{M})$
\end{algorithmic}
\end{algorithm}
The general architecture of the model is shown in \autoref{fig:Architecture}. The framework integrates both molecular graph structure and fingerprint information to comprehensively capture the properties of molecules. Specifically, for the molecular graph branch, a Graph Attention Network (GAT) is employed to extract local structural information, such as function groups, by emphasizing the importance of neighboring atoms. In parallel, a modified Graph Transformer module is introduced to capture global dependencies across the entire molecular graph, enabling the model to learn long-range interactions between atoms. The output of the GAT and Graph Transformer modules are then adaptively fused through a learned weighting mechanism, allowing the model to dynamically balance local and global information. This fused graph-based representation is subsequently combined with molecular fingerprint features through a Cross-Attention layer, which facilitates deep interaction between the two modalities. The resulting joint representation is used for downstream molecular property prediction tasks.
\subsection{Molecular fingerprints network}
To capture comprehensive molecular-level information, we constructed a composite fingerprint representation by concatenating three widely used molecular descriptors: Morgan fingerprint \cite{rogers2010extended} $M_{i}$, PubChem Fingerprint \cite{bolton2008pubchem} $P_{i}$, and Pharmacophore ErG Fingerprint \cite{stiefl2006erg} $E_{i}$. The concatenation of Morgan, PubChem, and Pharmacophore ErG fingerprints provides a complementary representation of molecular structures by capturing distinct yet synergistic aspects of chemical information. Specifically, the Morgan fingerprint encodes circular substructures and topological environments, the PubChem fingerprint reflects the presence of predefined substructure keys and function groups, while the Pharmacophore ErG fingerprint emphasizes 3D pharmacophoric patterns and spatial relationships relevant to bioactivity, thereby collectively enhancing the molecular representation for predictive modeling. Through the smiles $x_{i}$ of the molecules, we can get the corresponding molecular fingerprints. The Morgan Fingerprint encodes circular substructures, PubChem Fingerprint incorporates a variety of predefined structural keys, and ErG Fingerprint reflects pharmacophore features relevant for ligand-receptor interactions. By combining these fingerprints, we form a unified vector representation $U_{i}$ that encompasses a broad spectrum of molecular characteristics,
\begin{equation}
    U_{i}=[M_{i}\text{ }||\text{ }P_{i}\text{ }||\text{ }E_{i}].
\end{equation}
This concatenated fingerprint is then put into a multi-layer perceptron(MLP), which transforms it into a fixed-size embedding representing the global features of the molecule,
\begin{equation}
    FP_{i}=MLP(U_{i}).
\end{equation}
This embedding is subsequently used in downstream prediction tasks within our neural network framework so that it can get global information and choose some information that is useful for predicting the molecular property.
\subsection{Molecular featurization}
\begin{table}[htbp]
\renewcommand{\arraystretch}{1.2}
\centering
\caption{Initial Atomic and Bond Features}
\label{tab:features}
\begin{tabular}{@{} l c >{\raggedright\arraybackslash}p{11cm} @{}}
\toprule
\rowcolor{gray!20}
\textbf{Atom Features} & \textbf{Size} & \multicolumn{1}{c}{\textbf{Description}} \\
\midrule
Atom Symbol         & 16 & [C, N, O, F, Si, Cl, As, Se, Br, Te, I, At, others] (one-hot) \\
Degree              & 6  & Number of atoms connected to the atom (one-hot) \\
Formal Charge       & 1  & Electrical charge (integer) \\
Radical Electrons   & 1  & Number of radical electrons (integer) \\
Hybridization       & 6  & [sp, sp\textsuperscript{2}, sp\textsuperscript{3}, sp\textsuperscript{3}d, sp\textsuperscript{3}d\textsuperscript{2}, other] (one-hot) \\
Aromaticity         & 1  & Whether the atom is part of an aromatic system (one-hot) \\
Hydrogens           & 5  & Number of hydrogens connected to the atom (one-hot) \\
Chirality Type      & 4  & The chirality type of the atom (one-hot) \\
Ring                & 1  & Whether the atom is in a ring (one-hot) \\
Ring Type           & 4  & [3, 4, 5, 6] (one-hot) \\
Atomic Mass         & 1  & Mass of the atom (integer) \\
Implicit Valence    & 7  & [0, 1, 2, 3, 4, 5, 6, 7] (one-hot) \\
Hydrogen Acceptor   & 1  & Whether the atom acts as a hydrogen acceptor (one-hot) \\
Hydrogen Donor      & 1  & Whether the atom acts as a hydrogen donor (one-hot) \\
Acidic              & 1  & Whether the atom is part of an acidic group (one-hot) \\
Basic               & 1  & Whether the atom is part of a basic group (one-hot) \\
\midrule
\rowcolor{gray!20}
\textbf{Bond Features} & \textbf{Size} & \multicolumn{1}{c}{\textbf{Description}} \\
\midrule
Bond Type           & 5 & [exists, single, double, triple, aromatic] (one-hot) \\
Conjugation         & 1 & Whether the bond is conjugated (one-hot) \\
Ring                & 1 & Whether the bond is in a ring (one-hot) \\
Stereo              & 6 & [0, 1, 2, 3, 4, 5] (one-hot) \\
\bottomrule
\end{tabular}
\end{table}
To featurize the nodes and edges in the molecular graph, we encode them into vectors derived from a set of well-established atomic and bond characteristics. In this work, we employ 16 types of atomic features and 4 types of bond features to effectively capture the chemical identity of atoms and their local chemical environment (see \autoref{tab:features}). All features are encoded using one-hot vectors, except for the atomic mass and formal charge, which are preserved as continuous numerical values. For atom features, since hydrogen atoms are not explicitly represented in SMILES strings, the implicit valence of atoms is used as a proxy to infer the presence of bonded hydrogens thereby preserving chemical completeness. Additionally, we incorporate whether an atom functions as a hydrogen bond donor or acceptor, and whether it is part of an acidic or basic function group, as these properties are critical for molecular interactions and biological activity. For bond features, stereo-chemical information is also included, as it reflects the spatial arrangement of atoms, which can significantly influence molecular properties and interactions. As a result, for each molecular graph $G_{i}=\{\mathcal{V}_{i},\mathcal{E}_{i}\}$, each node $v\in \mathcal{V}_{i}$ is represented by a 57-dimensional feature vector $A_{v}$, and each edge $e_{vu}\in \mathcal{E}_{i} $ is encoded with a 13-dimensional feature vector $B_{vu}$.
\subsection{Molecular graph Representation}
Building on the Attentive FP framework \cite{xiong2019pushing}, which employs graph attention networks (GAT\cite{velickovic2017graph}), our approach utilizes a similar graph attention mechanism to effectively encode molecular graphs. However, unlike Attentive FP, which leverages deep multi-layer attention to capture global dependencies across the molecule, our design intentionally employs a shallower attention stack. This enables the model to focus on local substructure patterns, which are often more informative for certain molecular properties(e.g., functional groups or local motifs). In particular, we apply only a few attention layers to preserve locality and reduce over-smoothing, allowing the model to capture neighborhood-level information more effectively. To formally describe our Graph Attention Layer, consider a molecular graph $G_{i}=\{\mathcal{V}_{i},\mathcal{E}_{i}\}$, where $\mathcal{V}_{i}$ and $\mathcal{E}_{i}$ denote the sets of atoms (nodes) and bonds(edges), respectively. The initial feature representations for atoms $A_{v}$ and bonds $B_{vu}$ are constructed using a set of domain-specific chemical descriptors, capturing both intrinsic and relational chemical properties.\par
Initially, each atomic feature vector $A_{V}$ is projected through a fully connected layer followed by a ReLU activation function to obtain the initial node embedding:
\begin{equation}
    h_{v}^{0}=ReLU(W_{fc1}\cdot A_{v}).
\end{equation}
To incorporate information from each neighboring atom $u\in N(v)$, we concatenate its atomic feature $A_{u}$ with the bond feature $B_{vu}$ yielding the neighbor representation:
\begin{equation}
    n_{u}=[A_{u}\textbf{ }||\textbf{ }B_{vu}].
\end{equation}
This neighbor representation is further processed through a fully connected layer with ReLU activation function:
\begin{equation}
    h_{u}^{0}=ReLU(W_{fc2}\cdot n_{u}).
\end{equation}
After computing all initial node embeddings, we proceed to apply the $i$-th Graph Attention Layer. For each pair of connected nodes $(v,u)$, we calculate the unnormalized attention coefficient using a shared linear transformation followed by a LeakyReLU activation:
\begin{equation}
    e_{vu}^{i-1}=LeakyReLU(W_{i}[h_{v}^{i-1}||h_{u}^{i-1}]).
\end{equation}
These attention coefficients are then normalized over all neighbors of node $v$ using the softmax function:
\begin{equation}
    a_{vu}^{i-1}=softmax(e_{vu})=\frac{exp(e_{vu}^{i-1})}{\sum_{u\in N(v)} exp(e_{vu}^{i-1})}.
\end{equation}
Here, $N(v)$ denotes the set of neighboring nodes of atom $v$. Using the computed attention weights, the aggregated neighborhood representation is given by:
\begin{equation}
    c_{v}^{i-1}=elu(\sum_{u\in N(v)} a_{vu}^{i-1}\cdot W\cdot h_{u}^{i-1}).
\end{equation}
To enhance the expressive capacity of the model, we update the node embedding using a GRU\cite{chung2014empirical}, which integrates the aggregated neighborhood information with the previous hidden state:
\begin{equation}
    h_{v}^{i}=GRU(C_{v}^{i-1},h_{v}^{i-1}).
\end{equation}
where GRU denotes the gated recurrent unit function. In our framework, the GRU acts not only as an update mechanism but also plays a critical role in refining the local representations obtained from the attention layer. While the attention mechanism enables each node to selectively aggregate information from its neighbors, the GRU further enhances this representation by recurrently integrating the newly aggregated neighborhood context with the node’s prior hidden state. This temporal integration allows the model to effectively retain and propagate locally relevant information across multiple layers, which is particularly beneficial for capturing subtle but chemically meaningful patterns in the local molecular environment. Consequently, the inclusion of the GRU improves the expressiveness and robustness of node representations, thereby enhancing performance in downstream tasks such as molecular property prediction.\par
While GAT effectively captures local structural motifs through neighborhood attention, modeling long-range dependencies requires a more global mechanism. To address this, we adopt a Transformer-based architecture augmented with structural priors derived from the graph topology. The detailed structure of our modified Transformer is illustrated in \autoref{fig:Architecture}(B).Specifically, we incorporate the graph's normalized adjacency matrix $A$ into the self-attention computation to inject explicit structural information. The attention operation is modified as follows:
\begin{equation}
\mathcal{A}_i = \left( \lambda_a \cdot softmax\left(\frac{Q_i K_i^T}{\sqrt{d_k}} \right) + \lambda_b \cdot A \right)V_i,
\end{equation}
where $Q_i, K_i$ and $V_i$ are query, key, and value matrices, respectively, and $\lambda_a, \lambda_b$ are learnable scalars balancing the learned attention and the structural prior.\par
In this formulation, the softmax term provides global context by enabling interactions between all node pairs, while the adjacency matrix $A$ reinforces the known connections from the molecular graph. This design ensures that the model captures both long-range dependencies and local structural fidelity. In particular, for molecules with sparse or branched topologies, the learned attention offers flexible contextualization, while $A$ maintains connectivity-based inductive bias. For densely connected molecules, $A$ further stabilizes attention by anchoring it to chemically meaningful neighborhoods.\par 
Additionally, we replace the conventional Layer Normalization used in Transformer blocks with a novel Dynamic Tanh(DyT) \cite{zhu2025transformers} layer. DyT is a recently proposed alternative to traditional normalization layers, introduced as a lightweight and effective element-wise nonlinearity. Unlike Layer Normalization, which requires computing per-token statistics to rescale activations, DyT operates via a simple learnable squashing function defined as:
\begin{equation}
    DyT(x)=\gamma \cdot tanh(\alpha x)+\beta.
\end{equation}
where $\alpha$ is a learnable scalar that dynamically scales the input, and $\gamma,\beta$ are learnable affine parameters similar to those in normalization layers. This formulation preserves the S-shaped activation squashing effect observed in LayerNorm outputs, particularly for extreme activation values, while significantly reducing computational overhead. We have done relevant experiments and put them in the Supporting Information. \par

After processing molecular graphs through both GAT and Transformer pathways, we obtain two complementary node-level feature streams: local representations ${H_{i}}$ from GAT layers, where each $H_i \in \mathbb{R}^{n \times d}$ corresponds to the $i$-th layer’s output for all $n$ atoms, and global representations $H_T \in \mathbb{R}^{n \times d}$ from the Transformer, encoding long-range dependencies across the molecular graph.\par

To integrate these two perspectives, we introduce a Mixed Information Layer that adaptively fuses local and global features. First, we perform inter-layer aggregation over the outputs of all GAT layers. This is motivated by the observation that different GAT layers capture different neighborhood sizes (receptive fields), and aggregating them provides a more expressive local representation. The average over all GAT layers is computed as:
\begin{equation}
Mean(H) = \frac{1}{n} \sum_{i=1}^{n} H_i
\end{equation}
where $n$ is the number of GAT layers and $H_i \in \mathbb{R}^{n \times d}$ is the node embedding from the $i$-th GAT layer. This produces a unified node-level representation reflecting multi-hop local information. This is then passed through a linear layer and a GeLU activation to yield the processed local feature stream:
\begin{equation}
F_A = GeLU(Linear(Mean(H)))
\end{equation}
Here, $F_A \in \mathbb{R}^{n \times d}$ is the transformed local representation. The Linear layer enables feature transformation, while GeLU introduces smooth non-linearity for better generalization.
In parallel, the output from the Transformer encoder $H_T$ is also non-linearly activated:
\begin{equation}
F_T = GeLU(H_T)
\end{equation}
This step ensures consistency in activation treatment across both local and global streams. $F_T \in \mathbb{R}^{n \times d}$ contains node-level global features learned via content-aware and adjacency-guided attention mechanisms.
We combine $F_A$ and $F_T$ using a learnable scalar gate $\alpha \in [0,1]$, trained end-to-end with the model. The fusion is defined as:
\begin{equation}
F_{Mix} = \alpha \cdot F_A + (1 - \alpha) \cdot F_T
\end{equation}
This equation ensures an adaptive balance between local substructure sensitivity ($F_A$) and global topological awareness ($F_T$), allowing the model to modulate emphasis depending on the molecular property being predicted. For instance, properties like reactivity may benefit more from local patterns, whereas solubility may require more global contexts.
To convert node-level features into a single molecular representation, we introduce a Virtual Super Node $H_v$, which serves as a global summarization anchor. It is initialized as the sum of all atomic embeddings from the fused representation:
\begin{equation}
H_v = \sum_{i=1}^{n} H_i
\end{equation}
Here, $H_i \in \mathbb{R}^d$ is the embedding of the $i$-th atom in $F_{Mix}$, and $H_v \in \mathbb{R}^d$ acts as a global placeholder for the molecular context. We then compute atom-to-supernode attention scores to allow the model to weigh atoms differently, based on their relevance to the final molecular property. For each atom $i$, we compute:
\begin{equation}
e_{vi} = LeakyReLU(W_1 [H_v || H_i])
\end{equation}
where $[H_v || H_i]$ denotes concatenation of the global super node and atomic embedding, $W_1 \in \mathbb{R}^{d \times 2d}$ is a learnable weight matrix, and $e_{vi} \in \mathbb{R}$ is an unnormalized scalar attention score reflecting the interaction between the $i$-th atom and the global context.
These scores are normalized via softmax to obtain attention weights:
\begin{equation}
a_{vi} = \frac{\exp(e_{vi})}{\sum_{j=1}^{n} \exp(e_{vj})}
\end{equation}
Using the attention weights $a_{vi}$, we aggregate the atom embeddings into a context vector $C_v$:
\begin{equation}
C_v = elu\left(\sum_{i=1}^{n} a_{vi} \cdot W_2 \cdot H_i \right)
\end{equation}
where $W_2 \in \mathbb{R}^{d \times d}$ is a learnable transformation matrix, and $elu$ provides smooth activation to stabilize training and enhance expressiveness. $C_v \in \mathbb{R}^d$ captures a weighted summary of atomic features, focused on the most informative atoms.
Finally, we update the virtual node $H_v$ using a GRU to integrate this attention-refined context:
\begin{equation}
F_v = GRU(C_v, H_v)
\end{equation}
The Virtual Super Node $F_{v}$ serves as a global, task-adaptive representation of the entire molecule. It is subsequently passed into a MLP for downstream prediction tasks. Importantly, the introduction of this virtual node is not merely for summarization, but plays a crucial role in further refining the fused local-global information obtained from Adaptive Mixture Layer. By attending to most relevant atomic embeddings via an attention mechanism and incorporating this context through a GRU-based update, the virtual node effectively filters and distills the most predictive aspects of the mixed features. This process enhances the model's capacity to generate expressive and discriminative molecular representations, ultimately improving its performance on various prediction tasks.
\subsection{Final Representation Fusion and Output Layer}
To obtain the final molecular representation for prediction, we design a fusion mechanism that effectively integrates both structural and fingerprint-based features. Specifically, the molecular graph is processed through multiple layers of extraction and filtering, capturing both global and local structural information. This process yields a comprehensive structural representation of the molecule.\par
In parallel, the molecular fingerprint, defined as a predefined feature vector that encodes molecular substructures, is transformed through MLP to generate a learned fingerprint representation. To enable rich interaction between these two modalities, we employ a cross-attention mechanism that jointly considers the structural and fingerprint-based representations. This module allows the model to adaptive focus on the most informative components from each source, effectively enhancing the representational capacity.\par
The fused output from the cross-attention module, containing integrated and refined information from both the molecular graph and fingerprint domains, is then passed through a final MLP. This layer performs the ultimate transformation required for the target prediction task producing the final prediction of the molecular property.
\section{Results and Discussion}
\subsection{Benchmark datasets}
In this study, we evaluated the performance of our model on 10 publicly available data sets from Wu et al\cite{wu2018moleculenet}. These include five classification datasets: BACE, BBBP, Tox21, SIDER, and ClinTox, as well as five regression datasets including ESOL, FreeSolv, PDBbind-C, PDBbind-R and Lipophilicity. The classification tasks primarily focus on the prediction of physiology or biophysics, with the evaluation based on the ROC-AUC metric. The regression tasks target physicochemical properties and are evaluated using RMSE. These datasets span diverse molecular properties and cover a broad spectrum of molecular structures, providing a comprehensive testbed for assessing the generalization capability of our model. More information about these datasets can be found in \autoref{tab:datasets}.
\begin{table}[htbp]
\centering
\small
\renewcommand{\arraystretch}{1.2}
\setlength{\tabcolsep}{4pt}
\resizebox{\textwidth}{!}{%
\begin{tabular}{lllllllp{4.0cm}}
\toprule
\textbf{Category} & \textbf{Dataset} & \textbf{Molecules} & \textbf{Tasks} & \textbf{Splitting} & \textbf{Task type} & \textbf{Metric} & \textbf{Description} \\
\midrule
\multirow{3}{*}{Biophysics} 
  & PDBbind-C   & 168   & 1 &Random &Regression & RMSE & Binding affinity\\
  & PDBbind-R   & 3040  & 1&Random & Regression & RMSE & Binding affinity \\
  & BACE  & 1513 &1 &Scaffold & Classification & ROC-AUC & Inhibitory activity\\
\midrule
\multirow{4}{*}{Physiology}
  & BBBP     &2053 & 1 & Scaffold & Classification & ROC-AUC & BBB permeability \\
  & SIDER    &1427 & 27 & Random & Classification & ROC-AUC & Drug-related side effect \\
  & Tox21    &7831 & 12 & Random  & Classification &  ROC-AUC & Compound toxicity \\
  & ClinTox  &1478 & 2 & Random   & Classification &  ROC-AUC & Clinical toxicity \\
\midrule
\multirow{3}{*}{Physical chemistry}
  & ESOL     &1128 &1   & Random& Regression       & RMSE    & Aqueous solubility \\
  & FreeSolv &642   &1  & Random & Regression       & RMSE    & Hydration Free energy \\
  & Lipophilicity&4200&1 & Random& Regression      & RMSE    & Compound LogD values \\
\bottomrule
\end{tabular}
} 
\caption{An overview of molecular property prediction benchmark dataset.}
\label{tab:datasets}
\end{table}
\subsection{Baselines models}
To evaluate the performance of our model, we considered two groups of baseline models. The first group consists of supervised learning methods, including MoleculeNet\cite{wu2018moleculenet}, D-MPNN\cite{yang2019analyzing}, ML-MPNN\cite{wang2022advanced}, FP-GNN\cite{cai2022fp}, MVGNN\cite{ma2022cross}, LineEvo\cite{ren2023enhancing}, HimGNN\cite{han2023himgnn}, AttentiveFP\cite{xiong2019pushing}, ResGAT\cite{nguyen2024resgat}. The second group is based on self-supervised learning approaches, including GROVER\cite{rong2020self}, MGSSL\cite{zhang2021motif}, HiMOL\cite{zang2023hierarchical}, MolTailor\cite{guo2024moltailor}, MolGraph-LarDo\cite{zhang2024molecular}, PremuNet\cite{zhang2024pre}, MolGT\cite{chen2025pretraining}. A brief introduction to the aforementioned baseline model is provided below.
\begin{itemize}
    \item \textbf{MoleculeNet\cite{wu2018moleculenet}} employs a diverse set of benchmark models for molecular property prediction, encompassing traditional machine learning algorithms such as Random Forest, as well as Graph-based approaches including GCN.
    \item \textbf{D-MPNN\cite{yang2019analyzing}} utilizes a directed edge-based message passing framework to more effectively capture information about chemical bonds, which enhances the performance of molecular property prediction.
    \item \textbf{ML-MPNN\cite{wang2022advanced}} enhances molecular property prediction by integrating sequence-level information in addition to conventional graph-level features
    \item \textbf{FP-GNN}\cite{cai2022fp} aggregates molecular fingerprint features and graph-based representations to facilitate molecular property prediction.
    \item \textbf{MVGNN}\cite{ma2022cross} introduces a novel Cross-Dependent Message Passing Scheme that facilitates mutual interaction between node and edge features, thereby improving the model's expressive power.
    \item \textbf{LineEvo}\cite{ren2023enhancing} proposes a new architectural component, the LineEvo Layer, which captures hierarchical granular features from molecular structures to enhance prediction performance.
    \item \textbf{HimGNN}\cite{han2023himgnn} introduces motif-level representations to extract structural information from molecular graphs, which are subsequently integrated with traditional graph features to improve the accuracy of molecular property prediction.
    \item \textbf{GROVER}\cite{rong2020self} proposes a novel Transformer-based architecture that jointly leverages context prediction and motif prediction tasks to pretrain on large-scale molecular graphs, thereby learning transferable representations for molecular property prediction.
    \item \textbf{MGSSL}\cite{zhang2021motif} enhances molecular property prediction by combining motif-level generative tasks with multi-level self-supervised pretraining on molecular graphs. 
    \item \textbf{HiMOL}\cite{zang2023hierarchical} integrates node-level, motif-level, and graph-level representations, leveraging a pretraining framework to enhance performance on molecular predictions tasks.
    \item \textbf{AttentiveFP}\cite{xiong2019pushing} incorporates a new graph attention mechanism to learn molecular representations directly from molecular graphs. 
    \item \textbf{ResGAT}\cite{nguyen2024resgat} introduces a residual graph attention network that integrates multi-head attention with residual connections to enhance message passing in molecular graphs.
    \item \textbf{MolTailor}\cite{guo2024moltailor} proposes a prompt-based framework that generates task-specific molecular representations by conditioning a pre-trained language model on natural language task descriptions.
    \item \textbf{MolGraph-LarDo}\cite{zhang2024molecular} presents a hybrid framework that combines large language models with lightweight, domain-specfic graph neural networks to enhance molecular representation learning.
    \item \textbf{PremuNet}\cite{zhang2024pre} introduces a pre-trained multi-representation fusion network that integrates graph-based, sequence-based, and descriptor-based molecular representations.
    \item \textbf{MolGT}\cite{chen2025pretraining} proposes a multimodal pretraining framework for molecular representation that integrates graph structure, SMILES sequences, and expert knowledge using a unified graph transformer architecture.
\end{itemize}
The performance metrics of the baseline models were obtained from existing literature. Specifically, for most datasets, we adopted the baseline results reported in HimGNN \cite{han2023himgnn}, while for the PDBbind datasets, results were referenced from FP-GNN \cite{cai2022fp}. For other baseline models, we used the results as reported in their original papers, including ResGAT\cite{nguyen2024resgat}, MolTailor\cite{guo2024moltailor}, MolGraph-LarDo\cite{zhang2024molecular}, PremuNet\cite{zhang2024pre}, and MolGT\cite{chen2025pretraining}. To ensure a fair comparison with these baseline models, we aligned certain experimental settings with those used in the corresponding studies.
\subsection{Implementation details}
We implemented our model using the RDKit and Pytorch, employing the Adam \cite{kingma2014adam} optimizer to update model parameters. Hyperparameter tuning was conducted using Bayesian Optimization \cite{snoek2012practical} to efficiently explore the search space. Specifically, we optimized seven key hyperparameters: (1) the number of Transformer layers $N$, (2) the number of attention heads $h$, (3) the dimensionality of each attention head $d_{k}$, (4) the output dimension of the GAT module $G_{out}$, (5) the dropout rate in the feedforward network $F_{d}$, (6) the dropout rate in the GAT module $G_{d}$, and (7) the dropout rate in the Transformer module $A_{d}$. The results of hyperparameter search are summarized in \autoref{tab:hyper_search}. For each hyperparameter configuration, the model was trained using 5 different random seeds, and the configuration with the highest average performance was selected as the final setting. To enhance the generalizability of the model across datasets, we conducted hyperparameter optimization on the Lipophilicity dataset, which features a chemically diverse set of molecules with a broad range of sizes and structural complexities. As a regression task with continuous-valued targets and rich molecular variation, Lipophilicity serves as a strong candidate for identifying robust hyperparameter configurations. The optimal settings derived from this dataset were then directly applied to other benchmark datasets without additional tuning. As shown in \autoref{tab:Classificaiton1}, \autoref{tab:regression1} and \autoref{tab:regression2}, the transferred configurations yielded consistently competitive results such as the Transformer depth, GAT output dimension, and dropout rates exhibit strong transferability when optimized on a representative and heterogeneous dataset. These findings indicate that Lipophilicity-based tuning can provide a reliable initialization for related molecular property prediction tasks, thereby reducing the computational overhead of dataset-specific hyperparameters searches. For data splitting, we followed the protocol adopted in Wu et al.\cite{wu2018moleculenet}. Specifically, MoleculeNet utilizes two primary splitting strategies: random splitting, which randomly partitions molecules into training, validation, and test sets, and scaffold splitting, which separates molecules based on their Bemis-Murcko scaffolds\cite{bemis1996properties} to assess model generalization to novel chemical structures. Following this convention, we split each dataset according to the MoleculeNet-recommended strategy using an 8:1:1 ratio for training, validation, and testing. The validation set was used to select the final model for the better generalization ability. For each dataset, we trained our model using 10 different random seeds. All models were trained on NVIDIA GeForce RTX 4090.
\begin{table}[H]
\centering
\small
\begin{tabular}{
    c
    c c c
    l
    l l l
    l
}
\toprule
Iter & \textbf{$N$} & \textbf{$h$} & $d_k$ & \textbf{$G_{out}$} & \textbf{$G_{d}$} & \textbf{$F_{d}$} & \textbf{$A_{d}$} & RMSE \\
\midrule
1 & 5 & 13 & 104 & 140 & 0.50 & 0.55 & 0.25 & {0.639 ± 0.018}\\
2 & 4 & 13 & 120 & 140 & 0.05 & 0.20 & 0.20 & {0.638 ± 0.017}\\
3 & 3 & 20 & 120 & 140 & 0.40 & 0.20 & 0.20 & {0.643 ± 0.030} \\
4 & 2 & 17 & 64  & 150 & 0.25 & 0.35 & 0.00 & {0.636 ± 0.015}\\
5 & 3 & 19 & 128 & 140 & 0.35 & 0.55 & 0.20 & {0.655 ± 0.018}\\
6 & 3 & 17 & 120 & 120 & 0.60 & 0.30 & 0.25 & {0.638 ± 0.024} \\
7 & 3 & 16 & 64  & 120 & 0.05 & 0.45 & 0.15 & {0.638 ± 0.019} \\
8 & 5 & 14 & 104 & 130 & 0.60 & 0.10 & 0.35 & {0.637 ± 0.032} \\
9 & 6 & 19 & 128 & 160 & 0.15 & 0.35 & 0.15 & {0.643 ± 0.013} \\
10& 5 & 18 & 96  & 150 & 0.25 & 0.00 & 0.55 & {0.658 ± 0.031}\\
11& 4 & 19 & 120 & 150 & 0.25 & 0.10 & 0.25 & {0.625 ± 0.024}\\
12& 5 & 19 & 80  & 120 & 0.55 & 0.10 & 0.55 & {0.647 ± 0.018} \\
13& 5 & 14 & 112 & 150 & 0.35 & 0.10 & 0.30 & {0.653 ± 0.030} \\
14& 3 & 12 & 96  & 130 & 0.35 & 0.15 & 0.05 & {0.639 ± 0.018}\\
15& 4 & 17 & 96  & 100 & 0.10 & 0.20 & 0.60 & {0.634 ± 0.024} \\
16& 3 & 16 & 112 & 110 & 0.15 & 0.35 & 0.05 & {0.639 ± 0.023} \\
\rowcolor{gray!20}
17& 3 & 19 & 96  & 110 & 0.50 & 0.05 & 0.50 & {0.610 ± 0.016} \\
18& 2 & 14 & 64  & 100 & 0.25 & 0.15 & 0.60 & {0.629 ± 0.018} \\
19& 2 & 15 & 64  & 100 & 0.05 & 0.40 & 0.60 & {0.618 ± 0.017} \\
20& 5 & 16 & 64  & 100 & 0.25 & 0.45 & 0.50 & {0.631 ± 0.030} \\
\bottomrule
\end{tabular}
\caption{Bayesian optimization for Lipophilicity prediction task. The gray-highlighted row indicates the best hyperparameter setting.}
\label{tab:hyper_search}
\end{table}
\subsection{Performance analysis}
To evaluate the effectiveness of our proposed model, we conduct a comprehensive comparison several strong baselines across a suite of benchmark datasets covering both classification and regression tasks. \autoref{tab:Classificaiton1} presents the mean and standard deviation of ROC-AUC scores on the classification benchmarks. The proposed model achieves the highest performance on two out of five datasets and ranks second on one, demonstrating its robustness and consistency across diverse molecular property prediction tasks. Similarly, \autoref{tab:regression1} and \autoref{tab:regression2} summarize the RMSE results for regression tasks. Our model consistently outperforms all baselines, highlighting its superior ability to capture intrinsic molecular features and the complex relationships among atoms and substructures. In summary, the experimental results on both classification and regression tasks demonstrate the strong generalization ability and effectiveness of our model. By leveraging multi-level structural information and a hierarchical representation learning framework, the proposed model is capable of capturing complex molecular features and dependencies, leading to consistently superior performance across a variety of molecular property prediction benchmarks. These findings validate the design of our approach and its potential for broad applicability in computational chemistry and drug discovery tasks.
\begin{table}[H]
\centering
\small
\begin{tabular}{lccccc}
\toprule
\textbf{Methods} & \textbf{BACE} & \textbf{BBBP} & \textbf{SIDER} & \textbf{Tox21} & \textbf{ClinTox} \\
\midrule
MoleculeNet & 0.802 ± 0.038 & 0.877 ± 0.036 & 0.593 ± 0.035 & 0.772 ± 0.041 & 0.855 ± 0.037\\
ML-MPNN & 0.819 ± 0.036 & 0.891 ± 0.013 & 0.609 ± 0.017 & 0.796 ± 0.021 & 0.865 ± 0.027 \\
D-MPNN & 0.823 ± 0.038 & 0.911 ± 0.048 & 0.610 ± 0.027 & 0.808 ± 0.023 & 0.879 ± 0.040 \\
MVGNN & 0.829 ± 0.034 & 0.914 ± 0.039 & 0.614 ± 0.029 & 0.811 ± 0.026 & 0.884 ± 0.053 \\
FP-GNN & 0.845 ± 0.028 & 0.910 ± 0.027 & 0.598 ± 0.014 & 0.803 ± 0.024 & 0.765 ± 0.038 \\
GROVER & 0.835 ± 0.044 & 0.911 ± 0.008 & 0.621 ± 0.006 & 0.806 ± 0.017 & 0.882 ± 0.013 \\
MGSSL & 0.805 ± 0.072 & 0.845 ± 0.011 & 0.591 ± 0.016 & 0.788 ± 0.022 & 0.877 ± 0.021 \\
LineEvo & 0.825 ± 0.034 & 0.905 ± 0.039 & 0.608 ± 0.011 & 0.835 ± 0.015 & 0.890 ± 0.036 \\
HiMol & 0.840 ± 0.072 & 0.865 ± 0.064 & 0.593 ± 0.022 & 0.815 ± 0.010 & 0.702 ± 0.092 \\
HimGNN & \underline{0.856 ± 0.034} & \underline{0.928 ± 0.027} & 0.642 ± 0.023 & 0.807 ± 0.017 & 0.917 ± 0.030 \\
AttentiveFP & 0.810 ± 0.050 & 0.881 ± 0.040 & 0.640 ± 0.060 & \textbf{0.860 ± 0.050} & 0.845 ± 0.020 \\
ResGAT & 0.822 ± 0.030 & 0.875 ± 0.040 & 0.630 ± 0.050 & \underline{0.839 ± 0.050} & 0.888 ± 0.010\\
MolTailor& -- & 0.811 ± 0.021 & -- & 0.806 ± 0.013 & \underline{0.923 ± 0.044}\\
MolGraph-LarDo & 0.830 ± 0.011 & 0.731 ± 0.015 & 0.609 ± 0.023 & -- & --  \\
PremuNet &0.843 ± 0.005 &0.733 ± 0.006  & 0.626 ± 0.008& 0.740 ± 0.012& \textbf{0.992 ± 0.002} \\
MolGT & 0.845 ± 0.001 & 0.737 ± 0.010 & \textbf{0.654 ± 0.004} & 0.758 ± 0.002 & 0.889 ± 0.004\\
MLFGNN & \textbf{0.900 ± 0.027}&\textbf{0.934 ± 0.014}& \underline{0.647 ± 0.016} & 0.818 ± 0.007&0.892 ± 0.033\\
\bottomrule
\end{tabular}
\caption{Performance on classification tasks in terms of ROC-AUC. Higher values indicate better performance. The best results are highlighted in \textbf{bold}, and the second-best results are \underline{underlined}. "--" means data unavailable.}
\label{tab:Classificaiton1}
\end{table}

\begin{table}[h]
\centering
\begin{tabular}{lccc}
\hline
\textbf{Methods}&\textbf{PDBbind-C} & \textbf{PDBbind-R} \\
\hline
D-MPNN  & 1.910 ± 0.000& \underline{1.338 ± 0.000} \\
FP-GNN  & \underline{1.876 ± 0.000} & 1.349 ± 0.000 \\
MLFGNN & \textbf{1.816 ± 0.074} & \textbf{1.330 ± 0.022 } \\
\hline
\end{tabular}
\caption{Performance on PDBbind datasets in terms of RMSE. Lower values indicate better performance. The best results are highlighted in \textbf{bold}}
\label{tab:regression1}
\end{table}
\begin{table}[H]
\centering
\small
\begin{tabular}{lccc}
\toprule
\textbf{Methods} & \textbf{ESOL} & \textbf{FreeSolv} & \textbf{Lipophilicity} \\
\midrule
MoleculeNet & 1.068 ± 0.057 &2.398 ± 0.259 & 0.712 ± 0.041\\
ML-MPNN & 0.905 ± 0.225 & 2.264 ± 0.112 & 0.649 ± 0.046 \\
D-MPNN & 0.972 ± 0.097 & 2.170 ± 0.536 & 0.652 ± 0.051 \\
MVGNN & 0.889 ± 0.152 & 2.115 ± 0.239 & 0.610 ± 0.059 \\ 
FP-GNN & 1.282 ± 0.332 & 2.492 ± 0.649 & 0.679 ± 0.152 \\
GROVER & 0.921 ± 0.092 & 2.026 ± 0.127 & 0.646 ± 0.021 \\
MGSSL & 1.935 ± 0.332 & 2.890 ± 0.837 & 0.951 ± 0.192 \\
LineEvo & 0.929 ± 0.137 & 2.254 ± 0.418 & 0.624 ± 0.011 \\
HiMol & 0.986 ± 0.231 & 2.731 ± 0.440 & 0.716 ± 0.254 \\
HimGNN & 0.870 ± 0.154 & 1.921 ± 0.474 & 0.632 ± 0.016 \\
AttentiveFP &1.522 ± 0.130 & 3.558 ± 0.720 & 1.123 ± 0.030 \\
ResGAT &0.812 ± 0.070 & 1.473 ± 0.310& 0.683 ± 0.020  \\
MolTailor & 0.723 ± 0.049 & 1.788 ± 0.326 & 0.810 ± 0.016  \\
MolGraph-LarDo &1.276 ± 0.053 & 2.356 ± 0.078& 0.748 ± 0.017 \\
PremuNet &0.851 ± 0.038 & 1.858 ± 0.205 & 0.730 ± 0.010  \\
MolGT & 0.839 ± 0.006& -- & 0.788 ± 0.013  \\
MLFGNN &\textbf{0.641 ± 0.055}& \textbf{1.409 ± 0.203} & \textbf{0.608 ± 0.013} \\
\bottomrule
\end{tabular}
\caption{Performance on regression tasks in terms of RMSE. Lower values indicate better performance. The best results are highlighted in \textbf{bold}, and the second-best results are \underline{underlined}. "--" means data unavailable.}
\label{tab:regression2}
\end{table}

\subsection{Ablation study}
\begin{figure}[H]
    \centering
    \includegraphics[width=\textwidth]{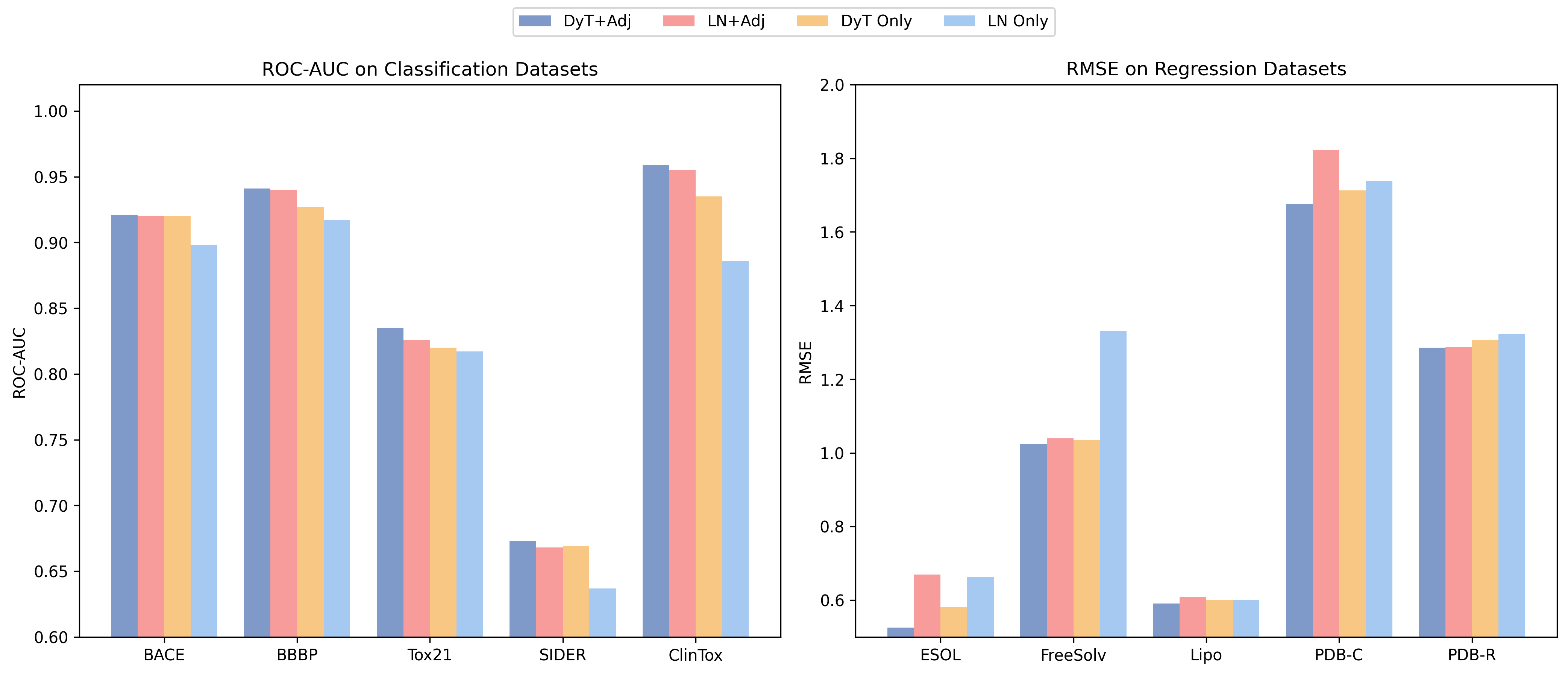}
    \caption{The performance of MLFGNN with different Graph Transformer architecture.}
    \label{fig:Ablation_Study_1}
\end{figure}

\begin{figure}[H]
    \centering
    \includegraphics[width=\textwidth]{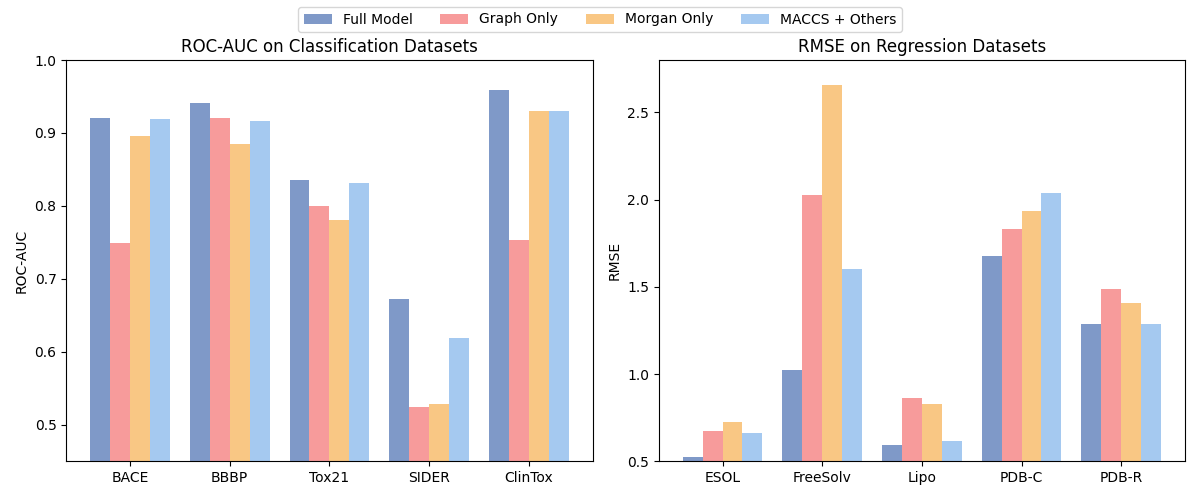}
    \caption{Ablation Study on Molecular Fingerprint Features.}
    \label{fig:Ablation_Study_2}
\end{figure}

\begin{figure}[H]
    \centering
    \includegraphics[width=\textwidth]{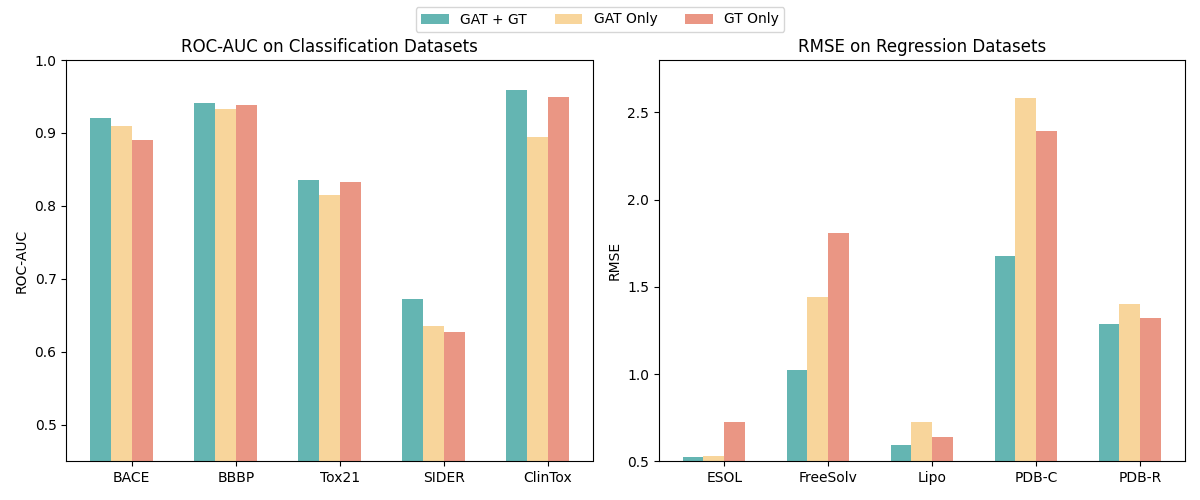}
    \caption{Ablation Study on Molecular Graph Module Structure.}
    \label{fig:Ablation_Study_3}
\end{figure}

In order to access the contribution of each component within our model architecture, we performed a series of ablation experiments targeting the Graph Transformer, the Molecular Graph representation, and the Fingerprint-based features. The ablation study results are presented in \autoref{fig:Ablation_Study_1}, \autoref{fig:Ablation_Study_2}, \autoref{fig:Ablation_Study_3}. Additionally, we conducted further ablation experiments to explore more detailed architectural variations. These supplementary results are provided in the Supporting Information for completeness and reference.\par
In \autoref{fig:Ablation_Study_1}, we investigate the impact of different configurations by replacing the original Graph Transformer module, which integrates DyT and Adjacency Matrix with three alternative variants: (1) Graph Transformer with only LN, (2) Transformer with only DyT, and (3) Transformer with LN and Adjacency Matrix. Experimental results across different datasets demonstrate that Transformer with DyT and Adjacency Matrix configuration consistently outperforms the others. These findings indicate that our proposed Transformer architecture is more effective in capturing inter-atomic relationships compared to other Transformer-based variants.\par
\autoref{fig:Ablation_Study_2} illustrates the results of the ablation experiments focusing on the Fingerprint module. We replaced the original concatenation of three molecular fingerprints in our model with two alternative configurations: using three single fingerprints and using the three fingerprints adopted in FP-GNN\cite{cai2022fp}(MACCS, PubChem, and ErG). Additionally, we performed an ablation in which the entire Fingerprint component was removed. The consistent superiority of the Morgan, PubChem, ErG fingerprints combination across all datasets demonstrates the effectiveness of combining diverse and complementary fingerprint features. In particular, the inclusion of Morgan fingerprints provides substantial gains over the MACCS-based combination used in prior models such as FP-GNN, suggesting that Morgan captures more informative substructural details. Moreover, the performance degradation observed when removing the Fingerprint component altogether underscores its importance in supplementing graph-based representation, enabling the model to leverage complementary molecular information for more accurate predictions.\par
\autoref{fig:Ablation_Study_3} presents the results of ablation experiments on the Molecular Graph representation. Specifically, we evaluated two configurations: one using only the GAT module and the other using the Graph Transformer module. The results show that combining GAT and Graph Transformer consistently outperforms using either module alone across all datasets. This suggests that our model effectively leverages the strengths of both components: GAT captures local substructural information such as molecular motifs and function groups, while the Graph Transformer captures global dependencies by modeling interactions between distant atoms. The complementary nature of local and global information enables the model to learn more expressive and task-relevant molecular representations.

\subsection{Visualization analysis}
\subsubsection{Analysis of Model Architecture Contributions}
\begin{figure}[H]
    \centering
    \includegraphics[width=\textwidth]{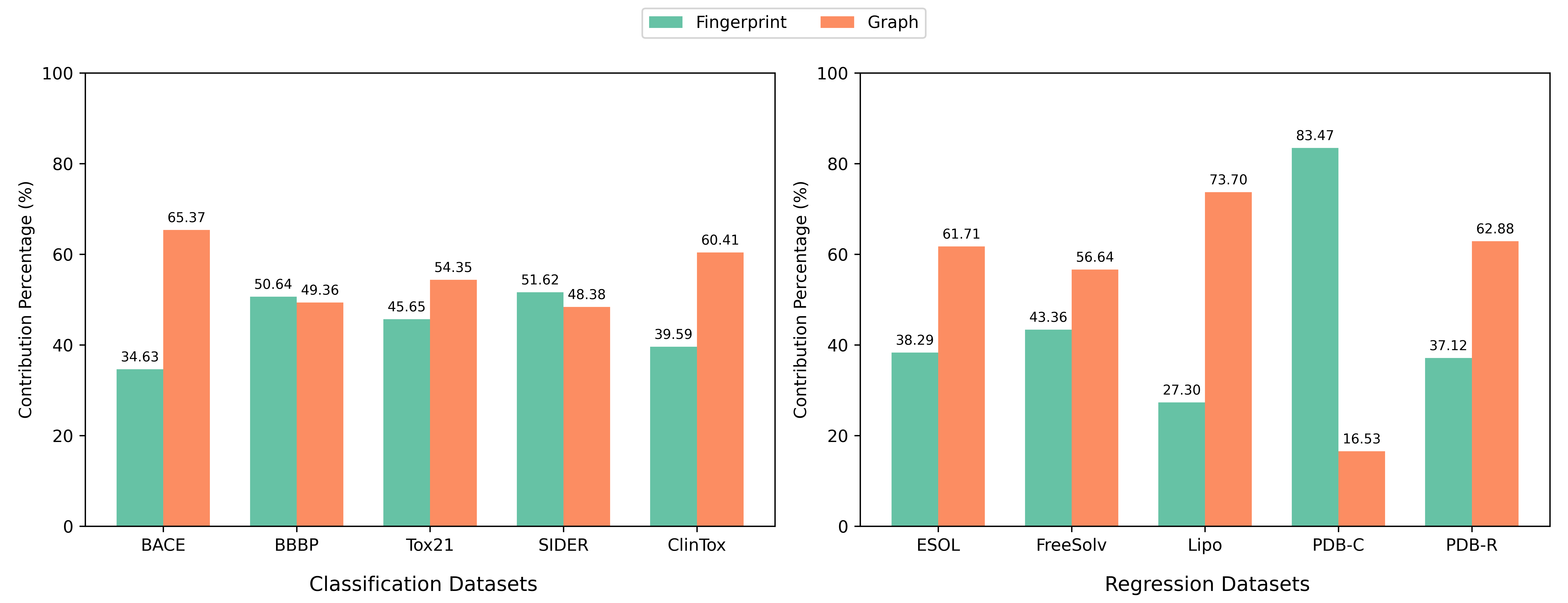}
    \caption{Contribution percentages of Fingerprint and Graph features across classification and regression datasets.}
    \label{fig:fingerprint vs graph}
\end{figure}

To gain a deeper understanding of the internal mechanisms of MLFGNN, we first analyzed the contributions of its architectural components through interpretability techniques and internal weight inspections. We employed SHAP (SHapley Additive exPlanation) values\cite{lundberg2017unified} to evaluate the relative contributions of the fingerprint-based and molecular graph-based representations. SHAP is a game-theoretic approach for explaining model outputs by attributing importance values to individual input features. By computing SHAP values separately for the fingerprint and graph components, we were able to quantify their respective roles in the final prediction. As illustrated in \autoref{fig:fingerprint vs graph}
, the contribution levels vary across datasets. Specifically,some tasks rely more heavily on molecular graphs, while others benefit more from fingerprints. This variation confirms that the two components provide complementary information, and their combination enhances the model’s generalization ability across diverse molecular property prediction tasks.\par
\begin{figure}[H]
    \centering
    \includegraphics[width=\textwidth]{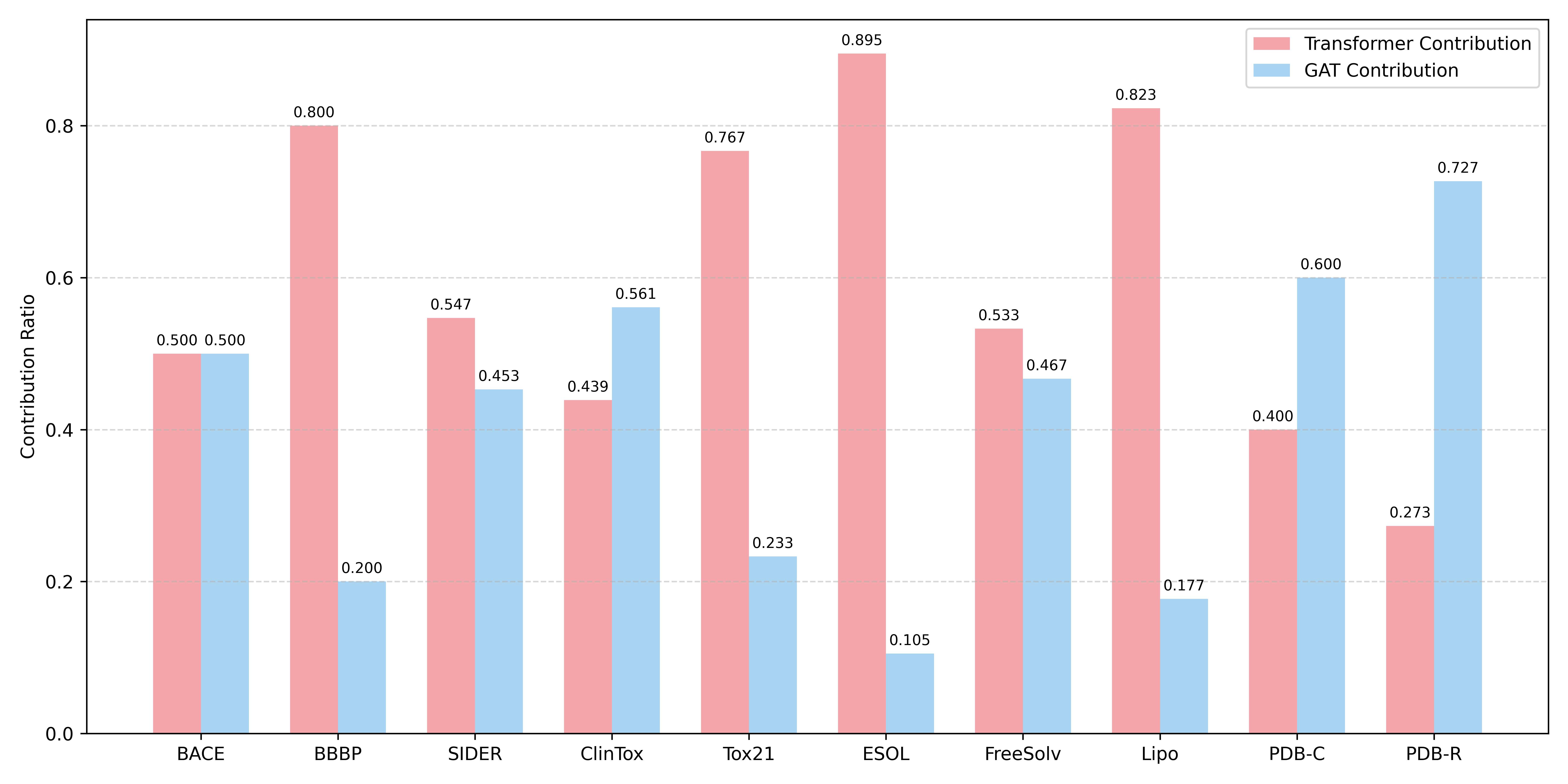}
    \caption{Comparison of Graph Transformer and GAT contributions across various datasets.}
    \label{fig:transformer vs gat}
\end{figure}

To further dissect the architectural design, we analyzed the Adaptive Mixture Layer, which integrates outputs from the Graph Attention Network (GAT) and the Graph Transformer layers. These two modules are responsible for capturing local and global structural information, respectively. By examining the learned mixing coefficients, we can assess the dynamic weighting between local and global features. As shown in \autoref{fig:transformer vs gat}, the model adaptively adjusts the relative contributions of GAT and Transformer layers based on dataset characteristics. This confirms the model's ability to extract task-specific information by balancing localized structural motifs with long-range dependencies, thus validating the necessity of incorporating both modules.
\subsubsection{Interpretation of Molecular Structures and Feature Attribution}
\begin{figure}[H]
    \centering
    \includegraphics[width=0.95\textwidth]{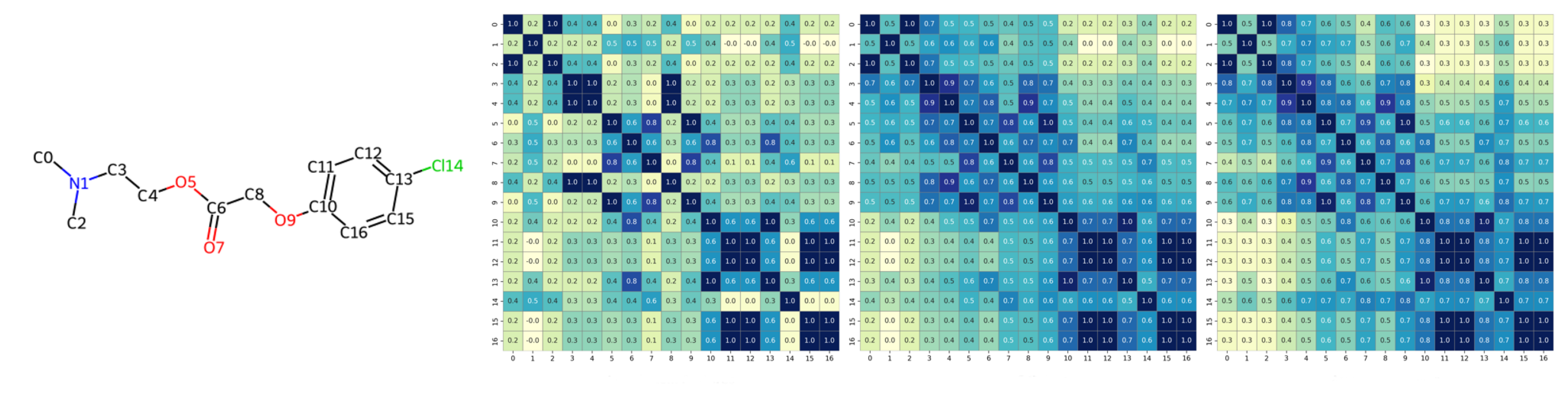}
    \caption*{(a) Visualization of atomic correlation patterns in the BBBP dataset.}
    
    \includegraphics[width=0.95\textwidth]{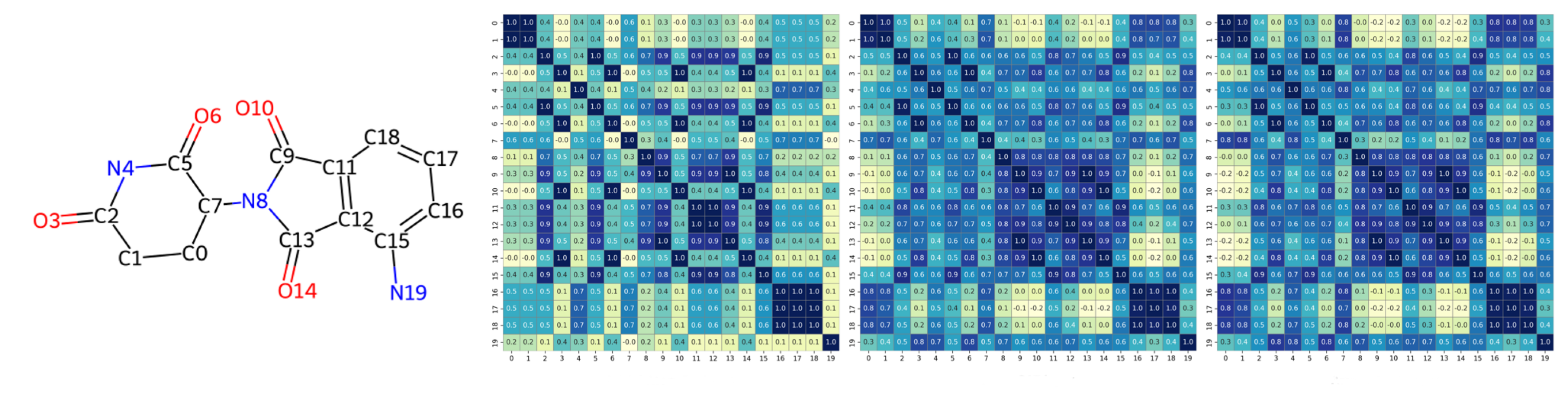}
    \caption*{(b) Visualization of atomic correlation patterns in the ClinTox dataset.}
    
    \includegraphics[width=0.95\textwidth]{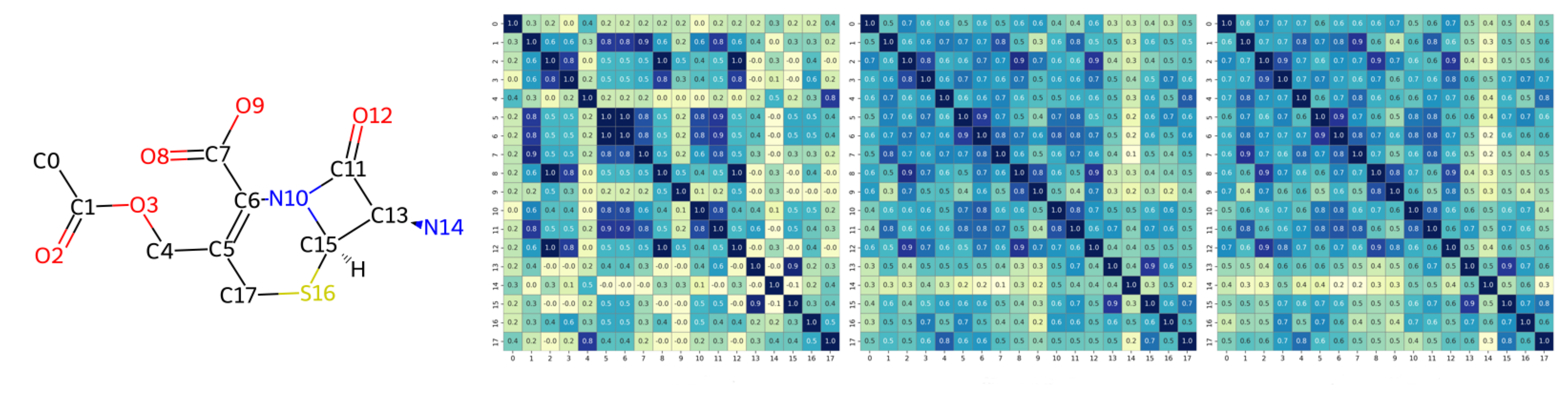}
    \caption*{(c) Visualization of atomic correlation patterns in the Tox21 dataset.}
    
    \caption{Molecular structures and atom feature heatmaps before and after GAT/mixed layers in some classification datasets. From left to right, each group of images shows: (1) the molecular structure, (2) the atom feature heatmap before training, (3) the atom feature heatmap after the GAT layer, and (4) the atom feature heatmap after the Mixed-information layer. (a) Visualization of atomic correlation patterns in the BBBP dataset. (b) Visualization of atomic correlation patterns in the ClinTox dataset. (c).Visualization of atomic correlation patterns in the Tox21 dataset.}
    \label{fig:classification_heat_map}
\end{figure}

\begin{figure}[H]
    \centering
    \includegraphics[width=0.95\textwidth]{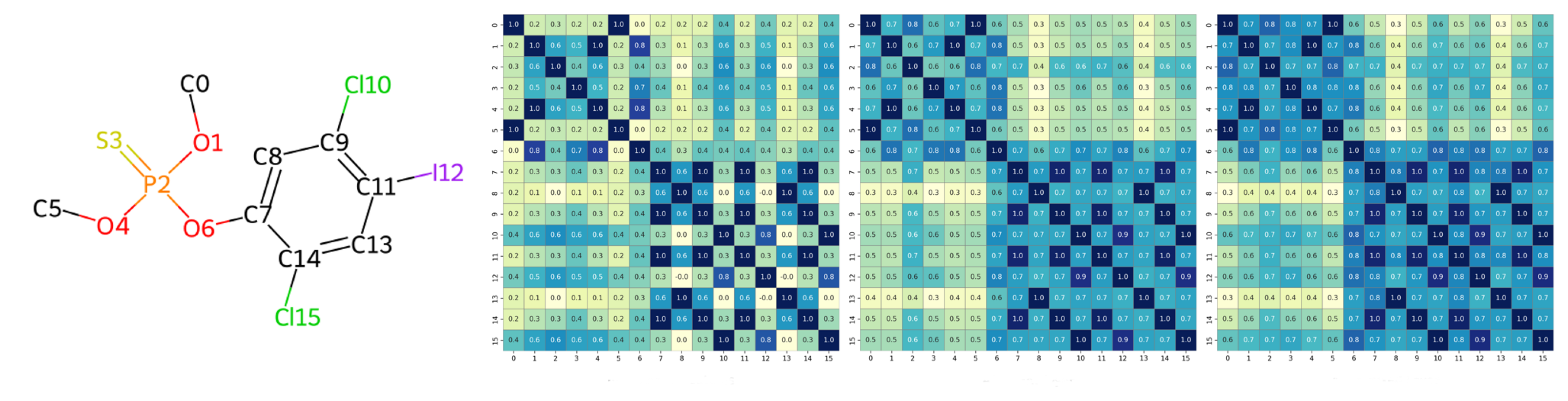}
    \caption*{(a) Visualization of atomic correlation patterns in the ESOL dataset.}
    
    \includegraphics[width=0.95\textwidth]{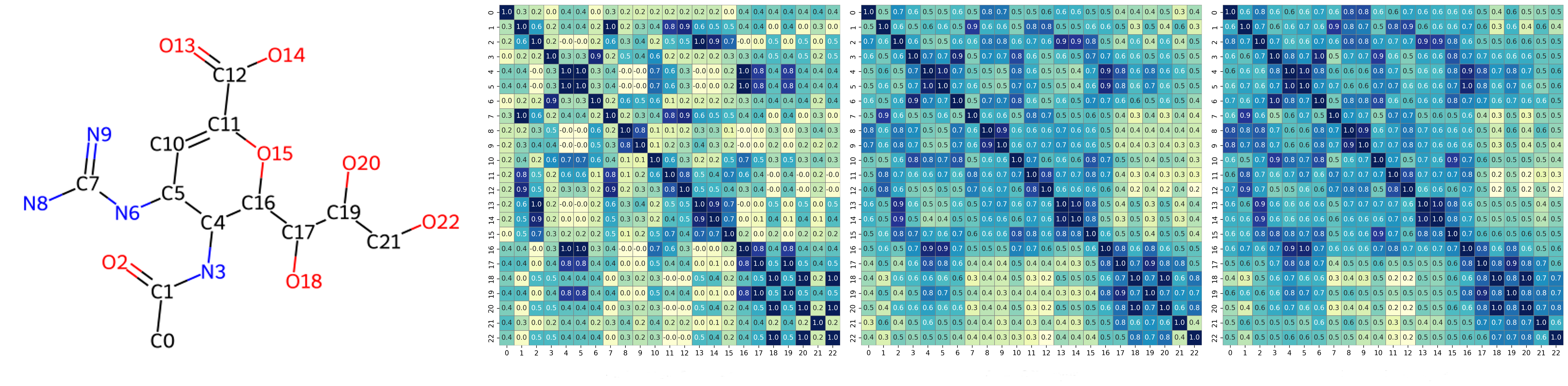}
    \caption*{(b) Visualization of atomic correlation patterns in the PDB-R dataset.}

    \includegraphics[width=0.95\textwidth]{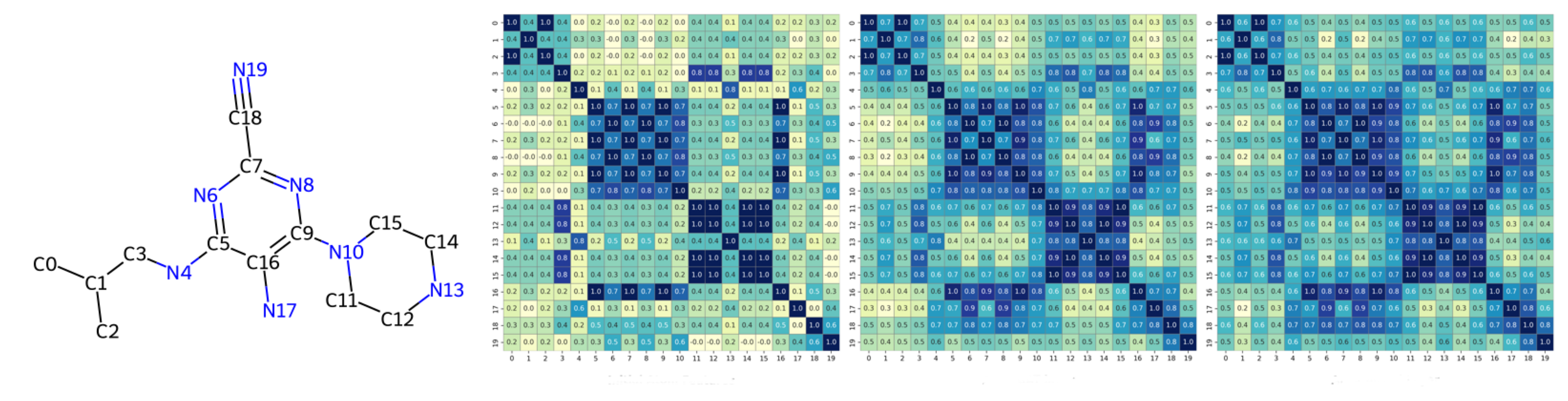}
    \caption*{(c) Visualization of atomic correlation patterns in the Lipo dataset.}
    
    \caption{Molecular structures and atom feature heatmaps before and after GAT/mixed layers in some regression datasets. From left to right, each group of images shows: (1) the molecular structure, (2) the atom feature heatmap before training, (3) the atom feature heatmap after the GAT layer, and (4) the atom feature heatmap after the Mixed-information layer. (a) Visualization of atomic correlation patterns in the ESOL dataset. (b) Visualization of atomic correlation patterns in the PDB-R dataset. (c) visualization of atomic correlation patterns in the Lipo dataset.}
    \label{fig:regression_heat_map}
\end{figure}

\autoref{fig:classification_heat_map} and \autoref{fig:regression_heat_map} illustrate the molecular structures alongside atom-level feature correlation heatmaps at different stages of MLFGNN. The visualizations highlight how atomic relationships evolve before training, after the GAT layer, and after the Mixed-information layer. The heatmaps reveal how different components of the model contribute to refinement of atomic feature representations. After passing through the GAT layer, localized structural patterns such as rings and specific functional groups, are more prominently captured, reflecting the GAT's strength in modeling local dependencies. When the Graph Transformer is further integrated  via the Adaptive Mixture Layer, the resulting representations incorporate additional long-range dependencies. This enables the model to better address challenges like activity cliffs, where subtle, spatially distant atomic differences significantly affect molecular properties. Overall, these visualizations demonstrate the complementary roles of local and global information in generating expressive molecular representations.\par
\begin{figure}[H]
    \centering
    \includegraphics[width=\textwidth]{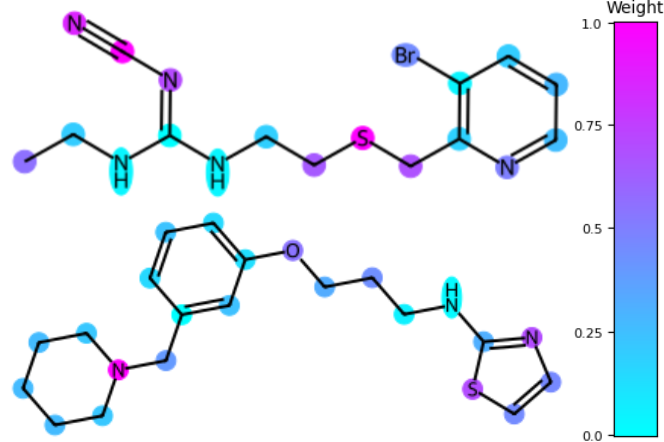}
    \caption{Visualization of atomic contributions to the molecular graph representation for two molecules from the BBBP dataset. The color of each atom reflects its importance in determining the final graph-level embedding, with the color scale ranging from low (blue) to high (pink). The molecule on the top has a BBB permeability label of 1 (permeable), while the one on the bottom is labeled 0 (non-permeable). This visualization highlights the regions of the molecules that contribute most significantly to the prediction of blood-brain barrier permeability.}
    \label{fig:BBBP_explain}
\end{figure}
\begin{figure}[H]
    \centering
    \includegraphics[width=\textwidth]{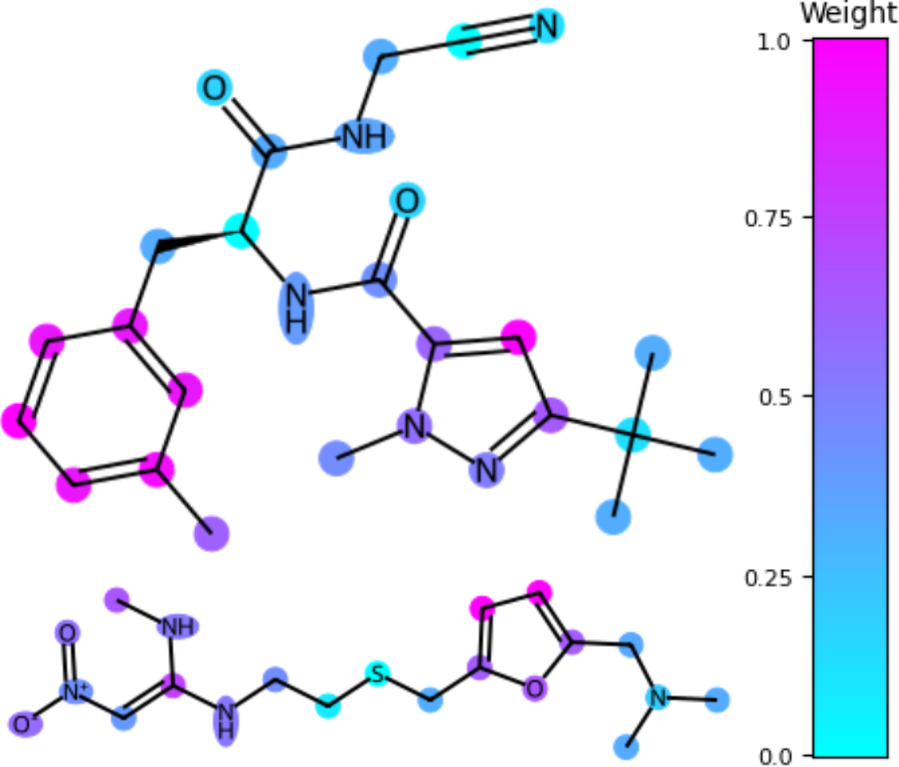}
    \caption{Visualization of atomic contributions to the molecular graph representation for two molecules from the Lipophilicity dataset. The color of each atom reflects its relative importance in forming the final molecular embedding, with pink indicating higher contribution and blue indicating lower. The molecule on the top has a measured lipophilicity value of 3.10, indicating strong lipophilicity, while the one on the bottom has a value of –0.95, indicating strong hydrophilicity. The visualization highlights the substructures that are more influential in determining the predicted lipophilicity.}
    \label{fig:Lipophilicity_explain}
\end{figure}

\autoref{fig:BBBP_explain} and \autoref{fig:Lipophilicity_explain} provide visualizations of atom-level contributions to molecular graph representations for two prediction tasks. In both cases, the model assigns higher importance to chemically meaningful substructures. For instance, in  \autoref{fig:BBBP_explain}, the highlighted atoms correspond to function groups, such as aromatic rings and nitrogen-containing motifs, which are known to influence BBB permeability. Similarly, in \autoref{fig:Lipophilicity_explain}, atoms within hydrophobic aromatic rings of polar functional groups receive higher attention, aligning well with known determinants of lipophilicity. These observations suggest that the model is capable of capturing domain-relevant chemical knowledge, supporting the interpretability and reliability of its learned representations.

\section{CONCLUSIONS}
In this paper, we present MLFGNN, a Multi-Level Fusion Graph Neural Network designed to enhance molecular property prediction by combining structural and multi-modal information. Our model addresses the limitations of conventional GNNs by integrating both local and global molecular features via a hybrid GAT–Graph Transformer architecture. Additionally, molecular fingerprints are incorporated through a cross-attention mechanism, enabling adaptive feature selection and improved alignment with downstream tasks. Experimental results across various benchmark datasets validate the effectiveness and generalizability of our approach. Moreover, interpretability analysis confirms that MLFGNN can identify chemically meaningful patterns, including functional groups and long-range dependencies, which are crucial for accurate prediction. Overall, our framework demonstrates the potential of combining intra-graph and inter-modal fusion strategies for robust and interpretable molecular representation learning.
\section{ASSOCIATED CONTENT}
\subsection{Data Availability Statement}
All the data sets and the source code of MLFGNN can be found at
\url{https://github.com/lhb0189/MLFGNN}. 
\section{AUTHOR INFORMATION}
\subsection{Corresponding Author}
\textbf{Hou-biao Li} --- School of Mathematical Sciences, University of Electronic Science and Technology of China, Chengdu 610054, China; \url{https://orcid.org/0000-0002-7268-307X}; Email: lihoubiao0189@163.com.\\[1ex]

\subsection{Authors}
\textbf{XiaYu Liu} --- School of Mathematical Sciences, University of Electronic Science and Technology of China, Chengdu 610054, China; \url{https://orcid.org/0009-0001-5333-0754}; Email: sumr020716@gmail.com.\par 
\textbf{Chao Fan} --- College of Management Science, Chengdu University of Technology, Chengdu 610059, China; \url{https://orcid.org/0000-0002-2111-1709}; Email: fanchao@cdut.edu.cn\par
\textbf{Yang Liu} --- School of Mathematical Sciences, University of Electronic Science and Technology of China, Chengdu 610054, China; \url{https://orcid.org/0009-0009-4912-4470} Email: lyy$\_$6161@163.com

\subsection{Author Contributions}
H.L. contributed to the study design and supervised the model construction and writing of this paper; X.L. contributed to the model conception, model construction, model training, and writing of the paper; C.F. contributed to the model construction, model training, and writing of the paper; Y.L. contributed to the model construction and writing of the paper; 
\subsection{Notes}
The authors declare no competing financial interest.
\section{ACKNOWLEDGMENTS}
The authors acknowledge the National Natural Science
Foundation of China (Grant: 11101071).
\section{Table of contents}
\begin{figure}[H]
    \centering
    \includegraphics[width=\textwidth]{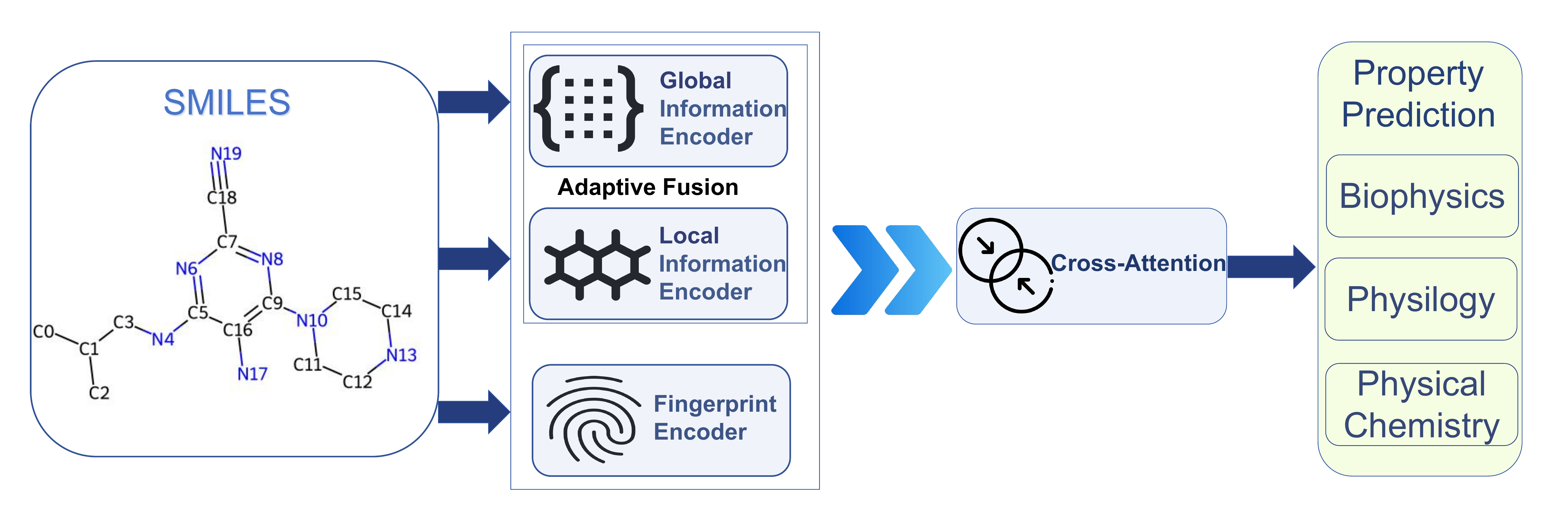}
    \caption{for table of contents use only.}
    \label{fig:table of Contents}
\end{figure}
\bibliography{achemso-demo}
\end{document}